*Brief Report*

# Challenges of 3D Surface Reconstruction in Capsule Endoscopy

**Olivier Rukundo** [1,2]

[1] Norwegian Colour and Visual Computing Laboratory, Department of Computer Science, Norwegian University of Science and Technology, Teknologiveien 22, 2815 Gjøvik, Norway; olivier.rukundo@meduniwien.ac.at
[2] Center for Clinical Research, University Clinic of Dentistry, Medical University of Vienna, Sensengasse 2a, 1090 Vienna, Austria

**Abstract:** Essential for improving the accuracy and reliability of bowel cancer screening, three-dimensional (3D) surface reconstruction using capsule endoscopy (CE) images remains challenging due to CE hardware and software limitations. This report generally focuses on challenges associated with 3D visualization and specifically investigates the impact of the indeterminate selection of the angle of the line–of–sight on 3D surfaces. Furthermore, it demonstrates that impact through 3D surfaces viewed at the same azimuth angles and different elevation angles of the line–of–sight. The report concludes that 3D printing of reconstructed 3D surfaces can potentially overcome line–of–sight indeterminate selection and 2D screen visual restriction-related errors.

## 1. Introduction

Capsule endoscopy (CE) is the newest and most patient-friendly endoscopic solution to gastrointestinal (GI) tract screening, particularly bowel cancer screening. To improve the CE–based screening process, the accurate and reliable evaluation of bowel pathologies can be facilitated by enhanced visualization of three-dimensional (3D) bowel surfaces. In CE, 3D visualization can be made possible by 3D reconstruction, which involves creating a 3D model of an object or scene from two-dimensional (2D) images or sensor data. One prominent approach for 3D reconstruction is the utilization of shape–from–shading algorithms [1,2]. Shape-from-shading algorithms, including Tsai's, Ciuti's, Barron's, and Torreao's, have been used to generate accurate 3D models [3]. Researchers have successfully applied shape–from–shading to represent the GI tract surface using 2D CE images [2]. Near-source perspective shape-from-shading enables precise 3D reconstructions of mucosal tissues [4]. Combining image stitching and shape–from–shading techniques generates comprehensive 3D maps [5]. Epipolar geometry enhances accuracy by constraining matching feature points for a more reliable 3D view [6].

However, despite these efforts, there are still several challenges and limitations that complicate the realization of 3D surface reconstruction in CE. For example, the CE hardware limitations and associated challenges make it infeasible to produce traditional 3D imaging, thus making 3D reconstruction from 2D images the only option in CE [7,8]. Specifically, operational and packaging–related challenges of the pill–cam or capsule endoscope [8,9] affect the traditional CE imaging procedure. On top of that, the GI environment is dark, and the natural peristalsis decides which lumen and mucosal surface to be imaged before viewing by gastroenterologists in a circular and monocular view [8,10], thus making the pathological evaluation efforts inaccurate or unreliable to some extent [3]. Another example is related to software limitations and associated challenges that make it difficult to accurately and reliably evaluate pathologies, such as user interface and interaction–based 3D visualization, the imprecise 3D mapping or inaccuracy of current techniques used for reconstruction of 3D surfaces from 2D images or frames [1–5,7–11].

In this report, the focus is on challenges associated with 3D visualization in CE. Specifically, the impact of the indeterminate selection of the angle of the line–of–sight for meaningfully visualizing the content of the reconstructed 3D surfaces from 2D images is demonstrated and discussed.

Figure 1a shows the line of sight in the 3D view context. This line starts at the center of the plot and points toward the camera or eye. As can be seen, two angles, the azimuth and the elevation, are the pillars of the line of sight. In this context, it can be understood that larger and noise–free images would be the key to achieving a better view of image objects' details before further processing. Therefore, preprocessing operations, such as image upscaling (via interpolation) and/or image filtering (via outlier removal), can help to leverage CE image quality in this direction. It is important to note that a particular emphasis was put on exposing and exploring the impact of the indeterminate selection of the angle of the line–of–sight for meaningfully visualizing the content of the reconstructed 3D surfaces from 2D images.

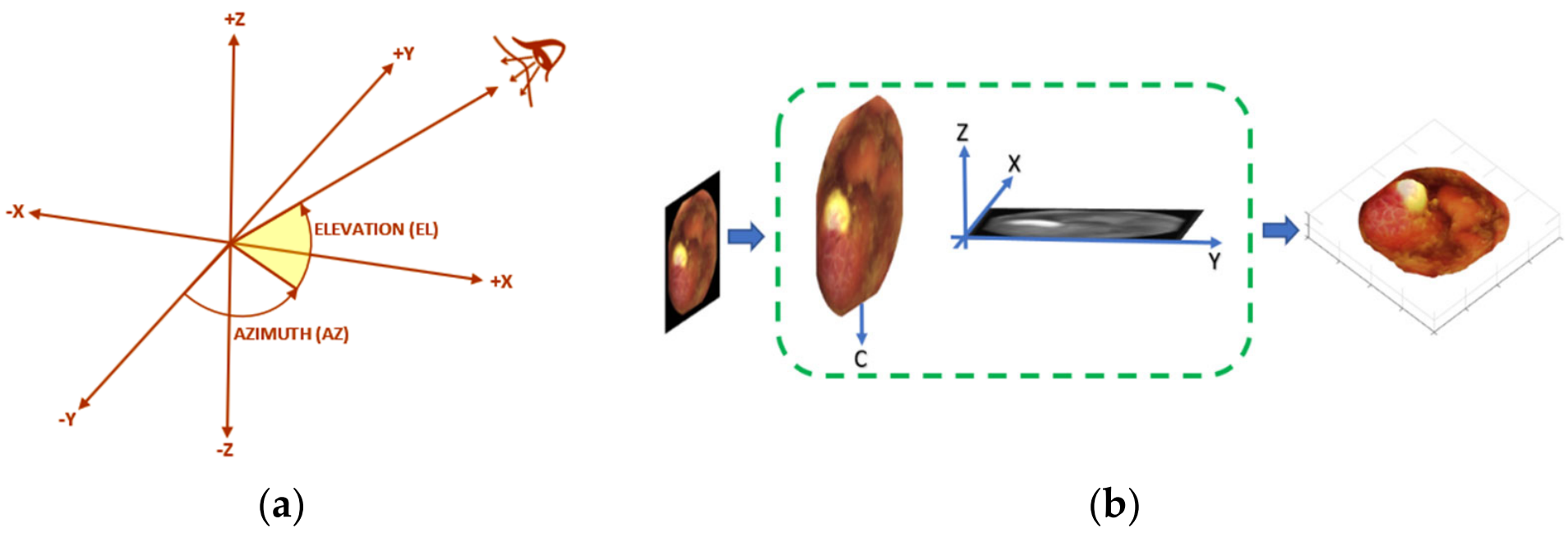

**Figure 1.** (**a**) Line–of–sight in the 3D view. (**b**) 3D surface reconstruction steps.

## 2. Materials and Methods

### *2.1. Pre-Processing*

Traditional pre–processing methods may include techniques for automatic processing or analysis purposes [12–14]. In this paper, we improved the quality of CE images for 3D surface reconstruction by preprocessing them with upscaling via interpolation and filtering via outlier removal.

#### 2.1.1. Interpolation

Interpolation is a widely used method in many fields to construct a new data value within the range of a set of known data [15–17]. This mathematical method pervades many applications in computer science and beyond. It enables us to obtain a high–resolution image from its low–resolution version [18]. In addition, image interpolation is practiced in improved definition television (IDTV) receiver design, photograph zooming and remote sensing [19]. Besides this, it is also applied in medical imaging, computer graphics, satellite imagery and in various other fields [15–23].

The author's prior studies generally demonstrated the performance of image interpolation algorithms in terms of effectiveness and efficiency [17,18]. Similary, other researchers demonstrated the effects of interpolation on the visual quality of digitally resized images [19–21].

In this work, the Lanczos interpolation method, referenced in [22], was used for image upscaling purposes. It is important to note that the Lanczos interpolation is based on the 3-lobed Lanczos window function as the interpolation function [22]. Given that Lanczos interpolation generally proved to lead to better outcomes than most interpolation methods, currently available in commercial software, it was therefore chosen over others to double the size of the input CE image before further processing.

#### 2.1.2. Filtering

Image filtering is the process of modifying an image to block or pass a particular set of frequency components [24]. There are many image–filtering techniques in the current literature, some of which have been specifically developed to remove outliers in digital images [24–27].

In this work, the simplest filtering procedure adopted includes rescaling image pixels and filtering using the 2D convolution kernel. Normally, the rescaling function scales the range of array elements to the desired interval. The desired interval is normally characterized by lower and upper bounds. The upper and lower bounds were determined using the mean and standard deviation of a given input CE image. The 2D convolution function was used with the convolution kernel size equal to 3 × 3 to filter the rescaled image. More details on 2D convolution using the kernel size 3 × 3 are provided in [28].

### *2.2. 3D Surface Reconstruction*

#### 2.2.1. Dataset

Our experimental dataset comprised five CE images (size 360 × 360 × 3) that were captured using the PillCam COLON. Note that these images were previously downloaded for our previous work, as presented in [29] from the capsule endoscopy database for medical decision support [30].

#### 2.2.2. Single Image 3D Reconstruction

MATLAB's 3D-colored surface function was used to plot the colored parametric surface defined by four matrix arguments X, Y, Z, and C. The lengths of interpolated images were used to create the row and column vectors needed by the meshgrid function to return the 2D grid coordinates, X and Y. The range of the Z argument was determined by the interpolated grayscale image, while the color scaling was determined by the range of C. Here, C was without the black background of the input image. This was achieved by first splitting the RGB color channels and extracting the mask, as well as computing its complement. The complement was separately added to each channel before concatenation. The shading model was determined by MATLAB's shading function. Figure 1b briefly illustrates the simplified 3D reconstruction steps from a single 2D CE image.

## 3. Results

Figure 2 shows three main columns, mainly (a), (b–c), and (d–e). The (a) column shows original CE images. Knowing whether these CE images contained bowel diseases was out of the scope of this work. Results presented in Figure 2 focused on demonstrating the need for determinate selection of the line–of–sight to better view 3D structures that contain these images. As can be seen, the (b) and (c) columns showed images that had 3D surfaces good and relevant enough to allow gastroenterologists to see the contents of 3D versions extracted from 2D CE images. However, the (d) and (e) columns showed images in which the structural contents were difficult to understand or find their relevance to the input images contents, showed in column (a). The reason for the lack of relevance of 3D surfaces was due to the elevation angle selected for images shown in columns (d) and (e). Here the EL = 0° while for columns (b) and (c), the EL = −80°. In both cases, the AZ = 0°. This demonstrated that, if not carefully selected, the angles of the line-of-sight could negatively affect the meaningfulness of the reconstructed 3D surfaces. In this context, a potential and promising solution would be to have reconstructed 3D surfaces printed in 3D objects to allow medical experts or gastroenterologists to directly observe them without 2D computer screen restrictions or related errors. Now, considering each of the five columns separately, it could be seen that (b) and (d) columns contained original or non-preprocessed images while (c) and (e) contained preprocessed images. Comparing the images in columns (b) and (c), as well as in columns (d) and (e), the images looked almost the same way (unless one zoomed in–and in such a case, it would be possible to notice differences in terms of smoothness of edges). In this way, the preprocessing did not significantly improve the quality of CE images, thus introducing the need for further research in this preprocessing direction.

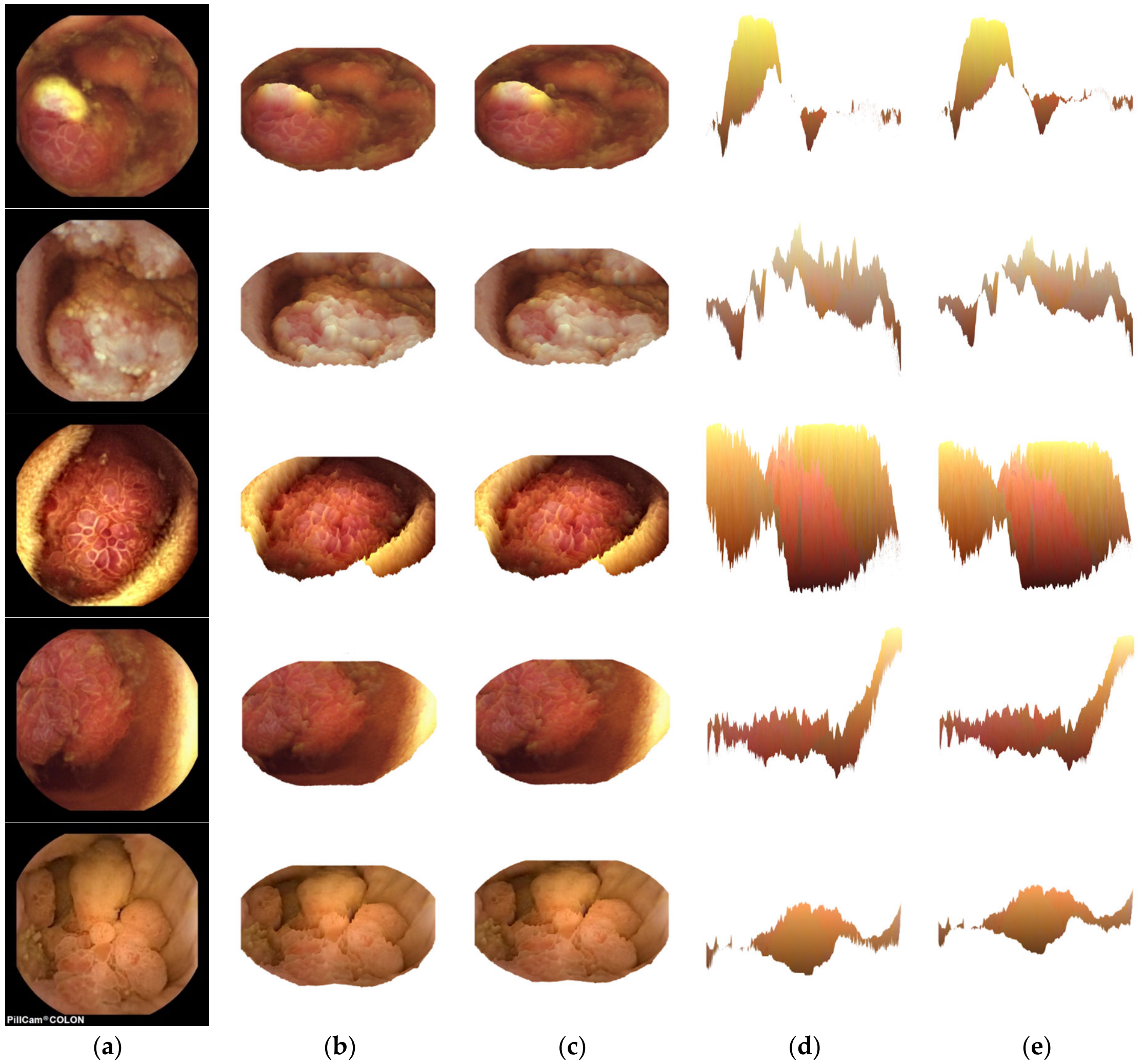

**Figure 2.** (**a**) CE image, (**b**) original AZ = 0°, EL = −80°, (**c**) preprocessed AZ = 0°, EL = −80°, (**d**) original AZ = 0°, EL = 0°, (**e**) preprocessed AZ = 0°, EL = 0°.

## 4. Discussions

The importance of 3D reconstruction in various aspects of capsule endoscopy imaging has been well–established, as reported by several works [1,2,4–10,16,31,32]. For example, these works highlight the benefits of 3D reconstruction in tasks such as characterizing subepithelial tumors [31], accurate measurements [32], enhanced lesion visualization [7], and promising results for polypoid structures and angioectasias [8]. In addition, according to authors in [2,6], the 3D reconstruction could provide clear surface recovery and improve the perception of the gastrointestinal (GI) tract. However, despite the significance of 3D reconstruction, there is a lack of focus on 3D software user interface-related challenges, such as 3D visualization-related, in existing works. Specifically, none of these works examined the effects of the irrelevance of 3D surfaces when the elevation angle and/or angle of the line of sight was selected indeterminately—which is why this report focused on demonstrating that the indeterminate selection of such an angle could negatively affect the meaningfulness of the reconstructed 3D surfaces. Although some endoscopists reported improved or non–improved visualization when referring to original 2D images in [7], the authors did not mention whether they encountered any challenges related to the 3D visualization of the reconstructed surfaces via user interfaces. Here, the author's main objective was to explore the accuracy of 3D reconstruction using innovative software and assess whether it led to enhanced lesion visualization in small bowel CE. In [8], authors noted the presence of highlights caused by lights reflected at various angles, which could potentially provide false information about the shape of the reconstructed surface. However, there was no further mention of the angle of view or the possibility that an indeterminate selection of the angle of the line of sight could lead to more highlights. This highlights the need for careful consideration of the angle of the line of sight or view when viewing reconstructed 3D surfaces, emphasizing the importance of a determinate selection of the angle, as demonstrated in this work. Another work [4] did not discuss the challenges related to the 3D software user interface or 3D visualization of reconstructed surfaces. Instead, the authors focused on other tasks involved in achieving 3D reconstructions of surfaces of interest. Similarly, in work [31], authors provided examples of 3D reconstructions from 2D images, but the view angles of these 3D surfaces were not defined or mentioned. Despite the lack of angle definition, the authors referred to other works to conclude that, at some percentage rate (less than 100%), the 3D versions presented enhanced visualization features compared to their 2D counterparts. This suggests that the lack of achieving 100% enhancement of visualization features can be attributed to the overlooking of 3D visualization challenges, particularly the failure to address the importance of a determinate line of sight. In [32], authors did not assess the effects of 3D visualization using the MiroCam MC4000 but instead evaluated its reliability in reconstructing 3D images and accurately calculating lesion size within a phantom model. The authors highlighted that the MiroCam MC4000 utilizes stereo-matching technology to enable the reconstruction of selected images in a 3D format for size calculation. They concluded that the estimated measurements highly correlated with the known sizes, showcasing the capabilities of this novel capsule. However, similar to previous cases, the authors overlooked the challenges of 3D visualization and instead focused on developing a method to reconstruct the 3D texture surface of the GI tract using a single CE image and the Shape from Shading technique [2]. In [6], authors acknowledged the need for a realistic and user-friendly 3D view to assist physicians in better viewing or observing the GI tract. However, they did not mention the visualization challenges associated with achieving this desired 3D view, particularly the importance of determining the angle of view or angle of the line of sight. In brief, the lack of work reporting on the 3D visualization challenges related to the 3D software user interface has led to the potential for demonstrating and reporting on these challenges. The impact of an indeterminate line–of–sight on 3D reconstructed surfaces has been evaluated. As a result, highlights the need for further consideration of the 3D software user interface challenges, particularly the angle of the line of sight, for optimal 3D visualization of capsule endoscopy imaging data.

## 5. Conclusions

In brief, this report sheds light on the specific challenge associated with meaningfully viewing the content of reconstructed 3D surfaces from CE. Preliminary results were presented to mainly demonstrate the extent to which the indeterminate selection of the line–of–sight could affect the 3D reconstruction-based analysis in CE. The report exposed and explored the potential to overcome the line–of–sight indeterminate selection challenge and suggested 3D printing of reconstructed 3D surfaces solution for the determinate selection of the line–of–sight and improving 3D visualization-based bowel cancer screening outcomes. Further research could extend this report's findings.

## 6. Future Perspectives of 3D Surface Reconstruction in CE

In brief, future perspectives can encompass exploring the potential of 3D printing, which would allow for determinate line–of–sight selection or leveraged 3D visualization, improving 3D user interfaces and visualization tools, and further investigating the impact of indeterminate line–of–sight on reconstructed surfaces. These three directions can contribute to enhancing the clinical utility and effectiveness of 3D reconstruction in CE.

**Funding:** This research was funded by the Research Council of Norway (Forskningsråd), Project number 260175, titled: Upscaling-Based Image Enhancement for Video Capsule Endoscopy, through Project number 247689, titled: Image Quality Enhancement in MEDical Diagnosis, Monitoring, and Treatment, IQ-MED.

**Institutional Review Board Statement:** Not applicable.

**Informed Consent Statement:** Not applicable.

**Data Availability Statement:** Data are available on our servers and can be shared upon request.

**Acknowledgments:** This research was supported by the Research Council of Norway. The author was affiliated with the Department of Computer Science, Norwegian University of Science and Technology, during the study. The author was affiliated with the University Clinic of Dentistry, Medical University of Vienna, during the revision and submission of the manuscript. The author would like to thank the editors and reviewers for their valuable comments.

**Conflicts of Interest:** The author declares no conflict of interest.

## References

1. Koulaouzidis, A.; Karargyris, A.; Giannakou, A.; Ang, Y.L.; Dabos, K.J.; Bartzis, L.; Bathgate, A.J.; Hayes, P.C.; Plevris, J.N. The Use of Three-Dimensional Reconstruction Software in Oesophageal Capsule Endoscopy: A Pilot Study from Edinburgh. *Glob. J. Gastroenterol. Hepatol.* **2014**, *2*, 84–91.
2. Zhao, Q.; Meng, M.Q. 3D Reconstruction of GI Tract Texture Surface using Capsule Endoscopy Images. In Proceedings of the 2012 IEEE International Conference on Automation and Logistics, Zhengzhou, China, 15–17 August 2012; pp. 277–282.
3. Koulaouzidis, A.; Iakovidis, D.K.; Karargyris, A.; Plevris, J.N. Optimizing Lesion Detection in Small Bowel Capsule Endoscopy: From Present Problems to Future Solutions. *Expert Rev. Gastroenterol. Hepatol.* **2015**, *9*, 217–235.
4. Prasath, V.B.S.; Figueiredo, I.N.; Figueiredo, P.N.; Palaniappan, K. Mucosal Region Detection and 3D Reconstruction in Wireless Capsule Endoscopy Videos using Active Contours. In Proceedings of the Annual International Conference of the IEEE Engineering in Medicine and Biology Society, San Diego, CA, USA, 28 August–1 September 2012; pp. 4014–4017.
5. Turan, M.; Pilavci, Y.Y.; Jamiruddin, R.; Araujo, H.; Konukoglu, E.; Sitti, M. A Fully Dense and Globally Consistent 3D map Reconstruction Approach for GI tract to Enhance Therapeutic Relevance of the Endoscopic Capsule Robot. *arXiv* **2017**, arXiv:1705.06524.
6. Fan, Y.; Meng, M.Q.; Li, B. 3D Reconstruction of Wireless Capsule Endoscopy Images. In Proceedings of the 2010 Annual International Conference of the IEEE Engineering in Medicine and Biology, Buenos Aires, Argentina, 31 August–4 September 2010; pp. 5149–5152.
7. Koulaouzidis, A.; Karargyris, A.; Rondonotti, E.; Plevris, J.N. PTU-021 3D Reconstruction in Capsule Endoscopy: A Feasibility Study. *Gut* **2013**, *62*, A50–A51.
8. Koulaouzidis, A.; Karargyris, A. Three-dimensional Image Reconstruction in Capsule Endoscopy. *World J. Gastroenterol.* **2012**, *18*, 4086–4090.
9. Kolar, A.; Romain, O.; Ayoub, J.; Faura, D.; Viateur, S.; Granado, B.; Graba, T. A System for an Accurate 3D Reconstruction in Video Endoscopy Capsule. *J. Embed. Syst.* **2009**, *2009*, 716317.
10. Ciuti, G.; Visentini-Scarzanella, M.; Dore, A.; Menciassi, A.; Dario, P.; Yang, G.Z. Intra-operative Monocular 3D Reconstruction for Image-guided Navigation in Active Locomotion Capsule Endoscopy. In Proceedings of the 4th IEEE RAS & EMBS International Conference on Biomedical Robotics and Biomechatronics (BioRob), Rome, Italy, 24–27 June 2012; pp. 768–774.
11. Prados, E.; Faugeras, O. Shape from Shading. In *Handbook of Mathematical Models in Computer Vision*; Springer: New York, NY, USA, 2006.
12. Münzer, B.; Schoeffmann, K.; Böszörmenyi, L. Content-based processing and analysis of endoscopic images and videos: A survey. *Multimed. Tools Appl.* **2018**, *77*, 1323–1362.
13. Padmavathi, G.; Shanmugapriya, D.; Kalaivani, M. Video preprocessing of image information for vehicle identification. *Int. J. Eng. Sci. Technol.* **2011**, *3*, 1526–1535. http://www.ijest.info/docs/IJEST11-03-02-280.pdf.
14. Hall, E.L.; Kruger, R.P.; Dwyer, S.J. A Survey of Pre-processing and Feature Extraction Techniques for Radiographic Images. *IEEE Trans. Comput.* **1971**, *20*, 1032–1044.

15. Karargyris, A.; Bourbakis, N. An Elastic Video Interpolation Methodology for Wireless Capsule Endoscopy Videos. In Proceedings of the 2010 IEEE International Conference on Bioinformatics and Bioengineering, Philadelphia, PA, USA, 31 May–3 June 2010; pp. 38–43.
16. Karargyris, A.; Bourbakis, N. Three-Dimensional Reconstruction of the Digestive Wall in Capsule Endoscopy Videos Using Elastic Video Interpolation. *IEEE Trans. Med. Imaging* **2011**, *30*, 957–971.
17. Rukundo, O. Evaluation of Rounding Functions in Nearest-Neighbour Interpolation. Int. J. Comput. Methods 2021, 18, 2150024.
18. Rukundo, O. Effects of Empty Bins on Image Upscaling in Capsule Endoscopy. In Proceedings of the SPIE 10420, Ninth International Conference on Digital Image Processing (ICDIP 2017), Hong Kong, China, 21 July 2017; Volume 10420.
19. Kumar, A.; Agarwal, N.; Bhadviya, J.; Tiwari, A.K. An efficient 2-D jacobian iteration modeling for image interpolation. In Proceedings of the 2012 19th IEEE International Conference on Electronics, Circuits, and Systems, Seville, Spain, 9–12 December 2012; pp. 977–980.
20. Pan, M.; Yang, X.; Tang, J. Research on interpolation methods in medical image processing. *J. Med. Syst.* **2010**, *36*, 777–807.
21. Luong, H.Q.; De Smet, P.; Philips, W. Image interpolation using constrained adaptive contrast enhancement techniques. In Proceedings of the IEEE International Conference on Image Processing ICIP, Genova, Italy, 14 September 2005; pp. 998–1001.
22. Rukundo, O. Effects of Image Size on Deep Learning. *Electronics* **2023**, *12*, 985.
23. Pal, R.; Begum, H.; Mukhopadhyay, S.; Sarkar, S.; Chakraborty, D.; Majumdar, S.; Sengupta, D. Edge Directed Radial Basis Function based Interpolation Towards PCA based PAN-sharpening. *Int. J. Remote Sens.* **2021**, *42*, 9047–9067.
24. Understanding Image Filtering Algorithms, Vision Systems Design. Available online: https://www.vision-systems.com (accessed on 16 March 2021).
25. Fan, L.; Zhang, F.; Fan, H.; Zhang, C. Brief review of image denoising techniques. *Vis. Comput. Ind. Biomed. Art* **2019**, *2*, 7.
26. Motwani, M.C.; Gadiya, M.C.; Motwani, R.C.; Harris, F.C. Survey of image denoising techniques. In Proceedings of the Proceedings of Global Signal Processing Expo and Conference (GSPx), Santa Clara, CA, USA, September 27-30, 2004.
27. Öktem, R.; Egiazarian, K.; Lukin, V.V.; Ponomarenko, N.N.; Tsymbal, O.V. Locally Adaptive DCT Filtering for Signal-Dependent Noise Removal. *EURASIP J. Adv. Signal Process.* **2007**, *2007*, 042472.
28. 2D Convolution in Image Processing, All About Circuits. Available online: https://www.allaboutcircuits.com/ (accessed on 16 March 2021).
29. Rukundo, O.; Pedersen, M.; Hovde, Ø. Advanced Image Enhancement Method for Distant Vessels and Structures in Capsule Endoscopy. *Comput. Math. Methods Med.* **2017**, *2017*, 9813165.
30. Koulaouzidis, A.; Iakovidis, D.K. KID: Koulaouzidis-Iakovidis Database for Capsule Endoscopy. Available online: http://is-innovation.eu/kid (accessed on 4 February 2016).
31. Nam, S.J.; Lim, Y.J.; Nam, J.H.; Lee, H.S.; Hwang, Y.; Park, J.; Chun, H.J. 3D Reconstruction of Small Bowel Lesions using Stereo Camera-based Capsule Endoscopy. *Sci. Rep.* **2020**, *10*, 6025.
32. Hawkes, E.; Keen, T.; Patel, P.; Rahman, I. PTH-024 Novel Capsule Endoscope with 3D Reconstruction and Lesions Size Calculation: First study with MiroCam MC4000. *Gut* **2019**, *68*, A25.